\pdfoutput=1

\documentclass[11pt]{article}

\usepackage[]{acl}

\usepackage{times}
\usepackage{latexsym}

\usepackage[T1]{fontenc}

\usepackage[utf8]{inputenc}

\usepackage{microtype}
\usepackage[ruled]{algorithm2e} 

\SetAlFnt{\small}
\SetAlCapFnt{\small}
\SetAlCapNameFnt{\small}
\SetAlCapHSkip{0pt}
\IncMargin{-\parindent}

\usepackage[normalem]{ulem}
\usepackage{amsmath}
\usepackage{graphicx}
\usepackage{mathtools}
\usepackage{caption}
\usepackage{subcaption}
\usepackage{mwe}
\usepackage{bm}
\usepackage{booktabs}
\usepackage{multirow}
\usepackage{nicefrac}
\usepackage{makecell}
\usepackage{paralist}
\usepackage{minitoc}
\usepackage[toc,page,header]{appendix}

\title{Digging Errors in NMT: Evaluating and Understanding Model Errors {from Partial Hypothesis Space}}

\author{%
  Jianhao Yan\thanks{~ Equal contribution.} \\
  WeChat AI, Tencent, China \\
  \texttt{elliottyan37@gmail.com} \\
  \And
  Chenming Wu$^{*}$ \\
  Tencent, China \\
  \texttt{wcm1994@gmail.com} \\
  \AND
  Fandong Meng \\
  WeChat AI, Tencent, China \\
  \And
  Jie Zhou \\
  WeChat AI, Tencent, China \\
}

\begin{document}
\maketitle
\doparttoc 
\faketableofcontents 

\begin{abstract}
Solid evaluation of neural machine translation (NMT) is key to its understanding and improvement. 
Current evaluation of an NMT system is usually built upon a heuristic decoding algorithm (e.g., beam search) and an evaluation metric assessing similarity between the translation and golden reference.
However, this system-level evaluation framework is limited by evaluating only one best hypothesis and search errors brought by heuristic decoding algorithms.
To better understand NMT models, we propose a novel evaluation protocol, which defines model errors with model's ranking capability over hypothesis space.
To tackle the problem of exponentially large space, we propose two approximation methods, top region evaluation along with an exact top-$k$ decoding algorithm, which finds top-ranked 
hypotheses in the whole hypothesis space, and Monte Carlo sampling evaluation, which simulates hypothesis space from a broader perspective. 
To quantify errors, we define our NMT model errors by measuring distance between  the hypothesis array ranked by the model and the ideally ranked hypothesis array.
After confirming the strong correlation with human judgment,
we apply our evaluation to various NMT benchmarks and model architectures.
We show that the state-of-the-art Transformer models face serious ranking issues and only perform at the random chance level in the top region. 
We further analyze model errors on architectures with different depths and widths, as well as different data-augmentation techniques, showing how these factors affect model errors. 
Finally, we connect model errors with the search algorithms and provide interesting findings of beam search inductive bias and correlation with Minimum Bayes Risk (MBR) decoding.
\end{abstract}

\section{Introduction}
Sequence-to-sequence models \citep{sutskever2014sequence, vaswani2017attention} have shown promising results in neural machine translation (NMT), where methods typically frame a conditional probability distribution from a source sentence to a target sentence. 
One key to the booming of neural machine translation is the sound evaluation, which shows the trajectory to a better model design and architecture.
The commonly used evaluation protocol of an NMT system comprises two main components: a search algorithm and an evaluation metric. The algorithm is responsible for decoding a translated sentence, and the metric computes the discrepancy between the generated translation and the reference. 


The above evaluation protocol is preferred as it is consistent with what we serve in production NMT.
It has an underlying assumption that the gap between an NMT model and the ideal model can be depicted by the gap between decoded translations and references.
However, this assumption does not always hold. 
Recent literature \cite{stahlberg-byrne-2019-nmt, meister-etal-2020-beam} points out that search errors brought by heuristic decoding methods would hide huge flaws of NMT models (model errors). The empty string is commonly scored with the highest probability among the model's probabilities over all hypotheses. 
Thus, disentanglement between search algorithms and NMT models is necessary for evaluating NMT systems.

Previous approaches disentangle search errors and model errors. However, they only take the \emph{mode}\footnote{Mode is the hypothesis with the highest probability in a distribution.} of the hypothesis space, i.e., all hypothesis accompanied with their probabilities, to evaluate model errors, which is not comprehensive. 
We ask two research questions: 
\begin{itemize}
    \setlength\itemsep{-1pt}
    \item \emph{Q1:How to define a more comprehensive evaluation over the hypothesis space?}
    \item \emph{Q2:With such evaluation, how do different architecture/data augmentation/search methods affect model errors?}
\end{itemize}

{To answer these questions, we introduce a new paradigm to evaluate model errors in hypothesis space.
The decoding and evaluation of model errors need to fit the requirements of the new paradigm. 
For the decoding algorithm, it should be both exact (not affected by search errors) and able to access more representative part of hypothesis space.
For the evaluation, it is essential to identify how good or bad these parts are quantitatively.}
Particularly, to deal with prohibitively large search space, we introduce two approximations: the top region evaluation, alongside with an exact top-$k$ decoding algorithm that not only avoids search errors but can access the top-ranked region of the whole hypothesis space, and the Monte Carlo sampling based evaluation. 
In addition, we provide formal definitions of evaluation in hypothesis space. We use hypothesis ranking (HR) as a proxy for measuring the distance between the model's hypothesis space and ideal hypothesis space. 


After confirming the strong correlation between our evaluation and human judgment, extensive experiments are conducted over three machine translation benchmarks with small, medium, and large sizes. We apply our proposed evaluation as a useful tool to analyze models and search algorithms. 
We identify that the state-of-the-art Transformer models have weak hypothesis ranking abilities only about the random chance level in the top region.
We further analyze model errors on models with different depths and widths, as well as applied with different data-augmentation techniques, showing how these affect model errors. 
In addition, we connect our model errors with search algorithms. 
Specifically, with our top-region evaluation, we provide quantitative results on beam search's lucky biases. With sampling-based evaluation, we show it correlates well with the promising minimum risk decoding.
\footnote{Codes will be released soon.}



Our contributions can be summarized as follows.
\begin{itemize}
    \item We propose an NMT model error evaluation over hypothesis space, with two approximated solutions addressing the prohibitively large hypothesis space and corresponding hypothesis-ranking (HR) metrics.
    \vspace{-5pt}
    \item We conduct in-depth analysis over various NMT techniques and find that the state-of-the-art Transformer models face severe hypothesis-ranking problems with abilities at the random chance level in top region. 
    \vspace{-5pt}
    \item We show that our evaluation is effective in analyzing the beam search's lucky biases and correlates well with the MBR decoding.
\end{itemize}

\addtolength{\abovedisplayskip}{-4pt}
\addtolength{\belowdisplayskip}{-4pt}
\section{Definitions}
We first introduce definitions of \emph{system level}, \emph{hypothesis mode} and \emph{hypothesis space evaluations}. 

\subsection{NMT Model and Hypothesis Space}
Give an NMT model $M$, a source sentence $x$ and a reference sentence $\hat{y}$. Most of the NMT models are auto-regressive models, which define a conditional distribution for a hypothesis $y_i$ as:
\begin{align}
    \label{eq:cand-prob}
    P(y_i|x) &= \prod_{t\in(1, T)} P(y_i^t | x; y_i^{1:t-1}), \nonumber \\
            & = M(x, y_i),
\end{align}
where $t$ represents the time step on target side and $T$ is the total length of $y_i$.

The hypothesis space of $M$ is defined as the set of all hypotheses given by $M$ along with their probabilities,
\begin{equation}
    \mathcal{Y} = \{(y_i, P(y_i|x)), ~\forall P(y_i|x)>0 \},
\end{equation}
and we refer to $\mathcal{Y}$ as $M$'s \textbf{hypothesis space}. \footnote{Note that there is a difference between the hypothesis space and search space, where the latter one illustrates the hypotheses that can be searched out.}
\subsection{System level Evaluation}
Given a decoding algorithm $F$ and an evaluation metric such as BLEURT or COMET \cite{sellam2020bleurt, rei2020comet}, the system-level evaluation of an NMT system usually proceeds by first decoding a hypothesis $y'$ from the hypothesis space:

\begin{equation}
    y' = F(\mathcal{Y}),
\end{equation}
where $F$ usually selects one or a few translation(s) with the highest step-by-step conditional probabilities from hypothesis space {according to the auto-regressive modeling}. 
Next, { system-level evaluation measures the similarity between $y'$ and reference $\hat{y}$}. 
\begin{equation}
    S_{\text{system}} = \text{Score}(\hat{y}, y').
\end{equation}

\subsection{Hypothesis Mode Evaluation}
It is recognized in previous literature \cite{niehues-etal-2017-analyzing, stahlberg-etal-2018-versatile, stahlberg-byrne-2019-nmt, meister-etal-2020-beam} that evaluating an NMT model and the decoding method as a whole system hinders the understanding of NMT model errors. Therefore, \citet{stahlberg-byrne-2019-nmt} propose an exact decoding method that finds the top-1 hypothesis $y_m$ over hypothesis distribution (mode) to evaluate model errors:
\begin{equation}
    y_m = \text{argmax}_{y \in \mathcal{Y}}(P(\mathcal{Y})),
    S_{\text{me}} = \text{Score}(\hat{y}, y_m).
\end{equation}
They find empty strings usually appear to be the modes of distributions and use the empty rate of modes to quantify the model errors.
We call this paradigm the mode-level evaluation in the following sections.

\subsection{Hypothesis Space Evaluation}
Selecting only one hypothesis in the whole hypothesis space loses much information of the hypothesis distribution and makes the evaluation biased.
Suppose we have two models A and B. Both of them have the mode hypotheses of empty string "<EOS>". 
However, other top hypotheses of A are high-quality translations, and those of B are low-quality translations. The mode-level evaluation will falsely regard them as the same.
To avoid such in-comprehensive bias, we define a new evaluation in the perspective of hypothesis space, which computes its distance with the ideal hypothesis space $\mathcal{Y}_{\text{ideal}}$:
\begin{gather}
    S_{\text{space}} = \text{D}(\mathcal{Y}, \mathcal{Y}_{\text{ideal}}).
\end{gather}

It is nontrivial to provide a sound definition to the ideal hypothesis space $\mathcal{Y}_\text{ideal}$ of an NMT model.
Here we mainly model one key attribute of the ideal space, which we call the \emph{hypothesis ranking ability}.
Intuitively, the ideal model's hypothesis space should align with the translation qualities over all hypotheses.  
In particular, if the translation quality of a specific hypothesis translation $y_i$ is better than that of $y_j$, the model's probability over $y_i$ should also be higher than that over $y_j$.
\begin{gather}
    P(y_i|x) > P(y_j|x) ~~\text{if}~~ Q(y_i) > Q(y_j) \nonumber \\
    \forall y_i, y_j \in \mathcal{Y},
\end{gather}
where $Q(y_i)$ is the translation quality function (e.g., COMET), and short for $Q(\hat{y}, y_i)$.

Hence, by extending such ability from pairwise to all hypotheses of a source sentence $x$, we define a proxy for ideal hypothesis space using the perfectly ordered hypothesis array of which the indices are sorted by translation quality. 
{
Formally, we define a perfect hypothesis-level ranking (HR) array $\bm{\mathcal{Y}_{\text{HR}}}$ over the hypothesis space $\mathcal{Y}$ with, 
\begin{gather}
    \label{eq:cand-rank}
    \bm{\mathcal{Y}_{\text{HR}}} = [y_{I_{\text{HR}}^0}, y_{I_{\text{HR}}^1}, \cdots, y_{I_{\text{HR}}^n}]; \\ 
    I_{\text{HR}} = \text{argsort}([Q(y_1), \cdots, Q(y_n)]).
\end{gather}
Analogously, we define $\bm{\mathcal{Y}_{\text{M}}}$ as the array sorted by model probabilities,
\begin{gather}
    \label{eq:model-rank}
    \bm{\mathcal{Y}_{\text{M}}} = [y_{I_{\text{M}}^0}, y_{I_{\text{M}}^1}, \cdots, y_{I_{\text{M}}^n}]; \\
    I_{\text{M}} = \text{argsort}([P(y_1|x), \cdots, P(y_n|x)]).
\end{gather}
Next, we can now define the model errors over hypothesis space with the distance between these two sorted arrays, 
\begin{gather}
\label{eq:me}
    S_{\text{dist}} = \text{D}(\bm{\mathcal{Y}_{\text{HR}}}, \bm{\mathcal{Y}_{\text{M}}}),
\end{gather}
where D is a certain distance function.
}




\addtolength{\abovedisplayskip}{-2pt}
\addtolength{\belowdisplayskip}{-2pt}

\section{Our Proposed Evaluation}
\label{sec:model_error}
{
{Two key designs of the evaluation over hypothesis space are the choice of distance functions and tackling the intractably large space. In this section, we first discuss our distance functions. Then, we propose two methods to simulate the hypothesis space, with the topmost and sampled hypotheses respectively.}

\subsection{Model Errors}
We propose two distance functions to describe ranking distance $D$ in this section. First, we propose an extended version of nDCG \citep{jarvelin2002cumulated}, 
which we coin \textbf{k}-approximated \textbf{R}anked \textbf{G}ains (\textbf{$\text{kRG}$}): 
\begin{gather}
    \label{eq:rank_nDCG}
    \text{kRG}(\bm{\mathcal{{Y}}_{\text{HR}}}, {\bm{{\mathcal{Y}}_{\text{M}}}}) = \frac{\text{DCG}_k({\bm{{\mathcal{Y}}_{\text{M}}}})}{\text{DCG}_k({\bm{{\mathcal{Y}}_{\text{HR}}}})}, \\
    \text{DCG}_{k}(\bm{\mathcal{Y}}) = \sum_{y_j \in \bm{\mathcal{Y}}} \frac{{f(y_j)}}{\log_2 (j+1)}, \\
    f(y_j) = k - \text{Rank}(y_j, \bm{\mathcal{{Y}_{\text{HR}}}}), 
\end{gather}
where $f(y_j)$ denotes the relevance score of a certain ranked hypothesis and $k$ is the length for approximated $\bm{\mathcal{Y}_{\text{HR}}}$ and $\bm{\mathcal{Y}_{\text{M}}}$.
$\text{kRG}$ directly measures the ranking of a model's hypotheses array, where 0 means a completely wrong ranking and 1 means a perfect ranking.

Next, in concern of translation quality of selected hypotheses, we further propose \textbf{k}-approximated \textbf{Q}uality-based \textbf{R}anked \textbf{G}ains (\textbf{$\text{kQRG}$}):
\begin{gather}
    \label{eq:brank}
    \text{kQRG}({\bm{\mathcal{Y}_{\text{HR}}}}, {\bm{{{\mathcal{Y}}}_{\text{M}}}}) = \frac{\text{DCG}_{qk}({\bm{\mathcal{Y}_{\text{M}}}})}{\text{DCG}_{qk}(\bm{\mathcal{Y}_{\text{HR}}})}, \\
    \text{DCG}_{qk}(\bm{\mathcal{Y}}) = \sum_{y_j \in \bm{\mathcal{Y}}} \frac{\text{Q}(y_j)}{\log_2 (j+1)},
\end{gather}
where we replace relevance score with translation quality $Q \in [0, 1]$ and normalize over $\bm{\mathcal{Y}_{\text{HR}}}$.
We approximate $\text{DCG}_{qk}(\bm{\mathcal{Y}_{\text{HR}}})$ with its upper-bound:
\begin{align}
    \text{DCG}_{qk}(\bm{\mathcal{Y}_{\text{HR}}}) &= \sum_{y_j \in \bm{\mathcal{Y}_{\text{HR}}}}  \frac{\text{Q}(y_j)}{\log_2 (j+1)}\\
    &<= \sum_{j\in[0:k]} \frac{1.0}{\log_2 (j+1)}.
\end{align}
$\text{kQRG}$ consider both how the hypotheses are ranked and whether these hypotheses have good translation qualities. 
Unlike kRG, the bound and interpretation of kQRG depends on the choice of translation quality functions, which we will discuss later. 

\subsection{Simulating Hypothesis Space}
As discussed above, it is intractable to obtain the HR array $\bm{\mathcal{Y}_{\text{HR}}}$ and model ranked array $\bm{\mathcal{Y}_{\text{M}}}$.
Our evaluation has to rely on approximations.
Here, we present two methods to approximate the hypothesis space, namely the top hypothesis region and Monte Carlo sampling. 

\subsubsection{Top Hypothesis Region}
\label{sec:topk-decoding}
While always being hindered by search errors, MAP decoding, the de facto standard search algorithm in NMT applications, seeks the topmost hypotheses from the whole space.
Thus, one reasonable approximation is to focus more on hypotheses with the highest probabilities, which are regarded, by the model, with great importance and are the globally optimal solutions for MAP decoding.
Formally, we define a top region model array:
\begin{gather}
    \label{eq:trunc}
    {\bm{\tilde{\mathcal{Y}}_{\text{M}}}} = \bm{\mathcal{Y}_{\text{M}}}[0:k];~ \tilde{I}_{\text{M}} = I_{\text{M}}[0:k],
\end{gather}
where $k$ denotes how many top-ranked hypotheses we consider. 

\vspace{3pt}\noindent\textbf{Exact Top-$k$ Decoding}
To find the topmost hypotheses, we extend the exact decoding algorithm \cite{stahlberg-byrne-2019-nmt} and propose a top-$k$ DFS-based exact decoding algorithm (Algorithm \ref{dfs_topk}).
Our decoding method is guaranteed to find the exact top-$k$ hypotheses from the model's hypothesis space.
Particularly, we traverse the search space of an NMT model in a depth-first manner. 
We enumerate all tokens in the vocabulary at each search step and concatenate them with the current history as the next possible translation prefixes. 
During the search process, we keep track of the current top-$k$ hypotheses that we find. 
Specifically, a minimum heap is used to maintain current top-$k$ hypotheses during the search procedure. 
The hypothesis with the lowest score in the minimum heap dynamically update our lower bound during searching:
Once we find a newly finished hypothesis (i.e., ended with </s>), we push the hypothesis into the heap and make adjustments to retain the heap size equals $k$. Then, we update the lower bound and truncate decoding paths. 
Finally, the hypotheses stayed in the minimum heap are returned.
We use beam search result as the initial bound of the search space and sort the vocabulary before enumeration for a faster update of lower bounds.
The implementation tricks and computational cost analysis can be found in Appendix \ref{sec:computation}.




\begin{algorithm}[t]

\LinesNumbered
\SetAlgoLined
\SetKwInOut{Input}{Input}
\SetKwInOut{Output}{Output}
\Input{x: Source sentence, y: Translation prefix (default: []),
p: $\log P(y|x)$ (default 0.0), k: Top-k hypotheses to output, V: Vocabulary.},  
\Output{List $l$ contains top-k hypotheses with log-probabilities.}

\SetKwFunction{FDFS}{dfsTopK}

\textbf{global} minHeap

\textbf{global $\gamma \gets -\inf$}
  
\SetKwProg{Fn}{Function}{:}{}
\Fn{\FDFS{$x$, $y$, $p$}}{
\If{$y[|y|-1] = </s> $}
{
    push(minHeap, $(p, y)$)
    
    \If{$\text{len}(\rm{minHeap}) > k$}{ pop(minHeap)}
    
    \If{$\text{len}(\rm{minHeap}) = k$}{$\gamma \gets \rm{minHeap[0][0]}$}
    
}

\For{$v \in V$}{
    
    $p' \gets p + \log P(v|x, y)$
    
    \If{$p' \geq \gamma$}{
        \FDFS{$x, [y;v], p' $} 

    }
}
\Return minHeap
}

\Return \FDFS{$x, [], 0.0 $} 
\caption{DFS-based Top-k Exact Search.}
\label{dfs_topk}
\end{algorithm}

\subsubsection{Hypothesis Region Sampling}
Besides the view of topmost region over the hypothesis space, we also provide a broad view for hypothesis space. We use Monte Carlo sampling to simulate the whole space as follows. Note that we slightly abuse the notation with $k$ as the number of samples.
\begin{gather}
    \label{eq:samp}
    y_i \sim P(y|x), i \in [0, k] \\
    {\bm{\tilde{\mathcal{Y}}_{\text{M}}}} = [y_{\tilde{I}_{\text{M}}^0}, y_{\tilde{I}_{\text{M}}^1}, \cdots, y_{\tilde{I}_{\text{M}}^{k
    }}], \\
    \tilde{I}_{\text{M}} = \text{argsort}([Q(y_1), \cdots, Q(y_{k})]).
\end{gather}
In both cases, there will be $k$ items in the array.

Then, we reorder hypotheses appeared in ${\bm{\tilde{\mathcal{Y}}_{\text{M}}}}$ to form a local HR array ${\bm{\tilde{\mathcal{Y}}_{\text{HR}}}}$,
\begin{gather}
    {\bm{\tilde{\mathcal{Y}}_{\text{HR}}}} = [y_{\tilde{I}_{\text{HR}}^0}, y_{\tilde{I}_{\text{HR}}^1}, \cdots, y_{\tilde{I}_{\text{HR}}^k}], \\
    \tilde{I}_{\text{HR}} = \text{argsort}([Q(y_{\tilde{I}_{\text{M}}^0}), \cdots, Q(y_{\tilde{I}_{\text{M}}^k})]).
\end{gather}

}






    
    
    


    
    




\section{Validation of Our Protocol}
\label{sec:valid}
This section validates the proposed protocol from the perspectives of translation quality, ranking capability and human evaluation.

\vspace{3pt}\noindent\textbf{Translation Quality.}
There are a number of sentence-level metrics proposed in neural machine translation. For example, there are string-based metrics like BLEU and ChrF \cite{papineni2002bleu,popovic2015chrf} and neural model-based metrics like BLEURT and COMET \cite{sellam2020bleurt, rei2020comet}. 
{
Recent studies and our human evaluation described later show that COMET scores are superior to other metrics in the correlations with human evaluation. \cite{kocmi2021ship, mathur2020results, freitag2021results}. 
Thus, we use COMET for main results of this paper.
}

\vspace{3pt}\noindent\textbf{Ranking Capability.}
The ranking capability of our protocol is evaluated by the nDCG metric, which is a widely used metric in many different areas that need to quantitatively measure the ranking efficacies \cite{liu-etal-2018-entity,agarwal-etal-2020-history}. The reliability of nDCG is well supported by previous literature. As a result, the validations of translation quality and ranking capability enable our protocol to be a solid evaluation protocol.

\vspace{3pt}\noindent\textbf{Human Evaluation.}
Moreover, we provide the human evaluation in this section to strengthen the validation of our protocol. We follow \cite{kocmi2021ship} to design the human evaluation. Specifically, we randomly select our NMT systems trained by the NIST Zh-En dataset into three evaluation groups. Each of which consists of comparison among three different systems, where we sample 50 sentences from NIST Zh-En test sets and provide top-$5$ exact decoding results ($\tilde{Y_M}$). As a result, each group has 750 sentences, and we conduct the human evaluation on a total of 9 systems.

We ask three professional Chinese-English translators to answer a question: how far are the array of translations from the perfect ranked outputs? (kQRG)
The annotators are required to give a score between 1 to 5. 
However, the scores are sometimes hard to give directly. Therefore, we ask human annotators to first have a sentence-level assessment of all translated sentences on a scale of [0, 100], following the source-based Direct Assessment method (DA, \citet{graham2017can}). 
We do not provide the reference to avoid the reference bias \cite{kocmi2021ship}. Then, annotators provide their ranking and total quality scores based on their scoring results of a system's top-$k$ (e.g., [40, 75, 40, 80]). 
{
We compute Pearson's/Spearman's Correlations between human scores and the corresponding kQRG on the top-$5$ translations. The results are 0.8554/0.8506 respectively \footnote{\url{}https://www.statstutor.ac.uk/resources/uploaded/spearmans.pdf}, which demonstrate a strong correlation between our proposed protocol and human judgments. 
We also conduct experiments comparing the correlation using different translation quality metrics other than COMET in the Appendix \ref{sec:human_corr}, including Sentence-BLEU, BLEURT \cite{sellam2020bleurt}, ChrF \cite{popovic2015chrf}, COMET-QE \cite{rei2020comet}, and COMET correlates well with human results.
We believe the above results validate our proposed protocol.
}




%
\section{Experiments and Findings}

In this section, we use our proposed evaluation protocol to evaluate two crucial factor of NMT systems -- model architecture and search algorithm. 

\vspace{3pt} \noindent \textbf{Setups.} All experiments are conducted over three commonly used NMT benchmarks, NIST Chinese-English, WMT'14 English-German, and WMT'14 English-French with small, medium, and large sizes.
The statistics of datasets, pre-processing and training details can be found in Appendix \ref{sec:exp_details}.





\vspace{3pt} \noindent \textbf{Evaluation Details.}
We use COMET \cite{rei2020comet} as our translation quality function among all experiments. 
We also provide results with ChrF \cite{popovic2015chrf} in the Appendix, as suggested in \citet{kocmi2021ship}.
By default, we use top-10 hypotheses for top region and 200 random samples for Monte Carlo sampling in all experiments. 

{
\vspace{3pt} \noindent \textbf{Interpretation.} We report kRG and kQRG results in our experiments. The kRG measures the `local' ranking ability of the top region of hypothesis space, directly representing whether the model correctly puts high-quality hypotheses over bad quality ones. The results range from 0 to 100\%, where 0 denotes a completely wrong ranking, and 100\% denotes a perfect ranking. Alternatively, kQRG measures two aspects: `local' ranking ability and hypothesis selection -- the quality of hypotheses that we can get from top-region or sampling. 
Using COMET trained with normalized z-scores, the kQRG values are not bounded by [0, 1] and may have negative values. 
A z-score above 0 means that the translation is better than average, and below 0 is the opposite. Thus, recall our definitions, we have two anchors to interpret kQRG values, where 0 means average translation qualities and 1 means perfect rankings with COMET values of 1, which is not the highest but a strong score. 

}



\begin{table}
\centering
\setlength\tabcolsep{2pt}
\begin{tabular}{l||c || c | c | c}
\toprule
\multirow{2}{*}{\bf Method} & \bf System &\textbf{{Mode}} & \bf Top & \bf Sample\\
\cmidrule{2-5}
& \textbf{BLEU} & \textbf{\# Emp} & \textbf{$\text{kQRG}$} & \textbf{$\text{kQRG}$} \\
\midrule
Transformer & 27.22 & 64.70 & -60.39 & -106.75 \\
\midrule
w/o LS & 26.76 & 34.85 &-17.75 & -25.07\\
w/ para BT & 27.36 & 27.26 & -13.04 & -19.52 \\
w/ para FT & 28.06 & \bf~~0.93 & \bf{~43.27} & \bf{~10.43} \\
\midrule
w/ 12-layer Enc & 27.75 & 58.11 & -50.57 & -104.94\\
w/ 18-layer Enc & 28.03 & 53.58 & -43.88 & -97.46 \\
\midrule
w/ Dim 768 & 28.00 & 50.18 & -43.33 & -101.23 \\
w/ Dim 1024 & \bf{28.49} & 44.72 & -34.56 & -84.93 \\
\bottomrule
\end{tabular}
\caption{Model errors of different models in WMT'14 En-De task. `para BT' and `para FT' denote back-translation and forward-translation over parallel golden data, and LS denotes label smoothing.}
\vspace{-20pt}
\label{table:ende}
\end{table}

\subsection{Findings on NMT Techniques}
\label{sec:find_techinique}
Table \ref{table:ende} demonstrates the results for different Transformer-based models in WMT'14 En-De. Results across different languages and other translation metrics can be found in Appendix \ref{sec:additional_results} and are consistent with our main results. We make following observations:

\vspace{3pt}\noindent\textbf{\emph{{1}. Failure of mode evaluation.}} 
Let us take a look at the empty rates, the evaluation for model errors proposed in previous literature. We find that removing label smoothing, adding pseudo-parallel data will drastically decrease the number of empty rates, even close to 0 (``\emph{para FT}''), indicating an almost perfect model with tiny model errors. However, it is not the case. Our kRG and kQRG results indicate that the model still has much to improve. 
These demonstrate that mode-level evaluation collapses when evaluating certain models and the superiority of our evaluation.

\begin{figure}
\centering
\begin{subfigure}{.5\linewidth}
  \centering
  \includegraphics[width=\linewidth]{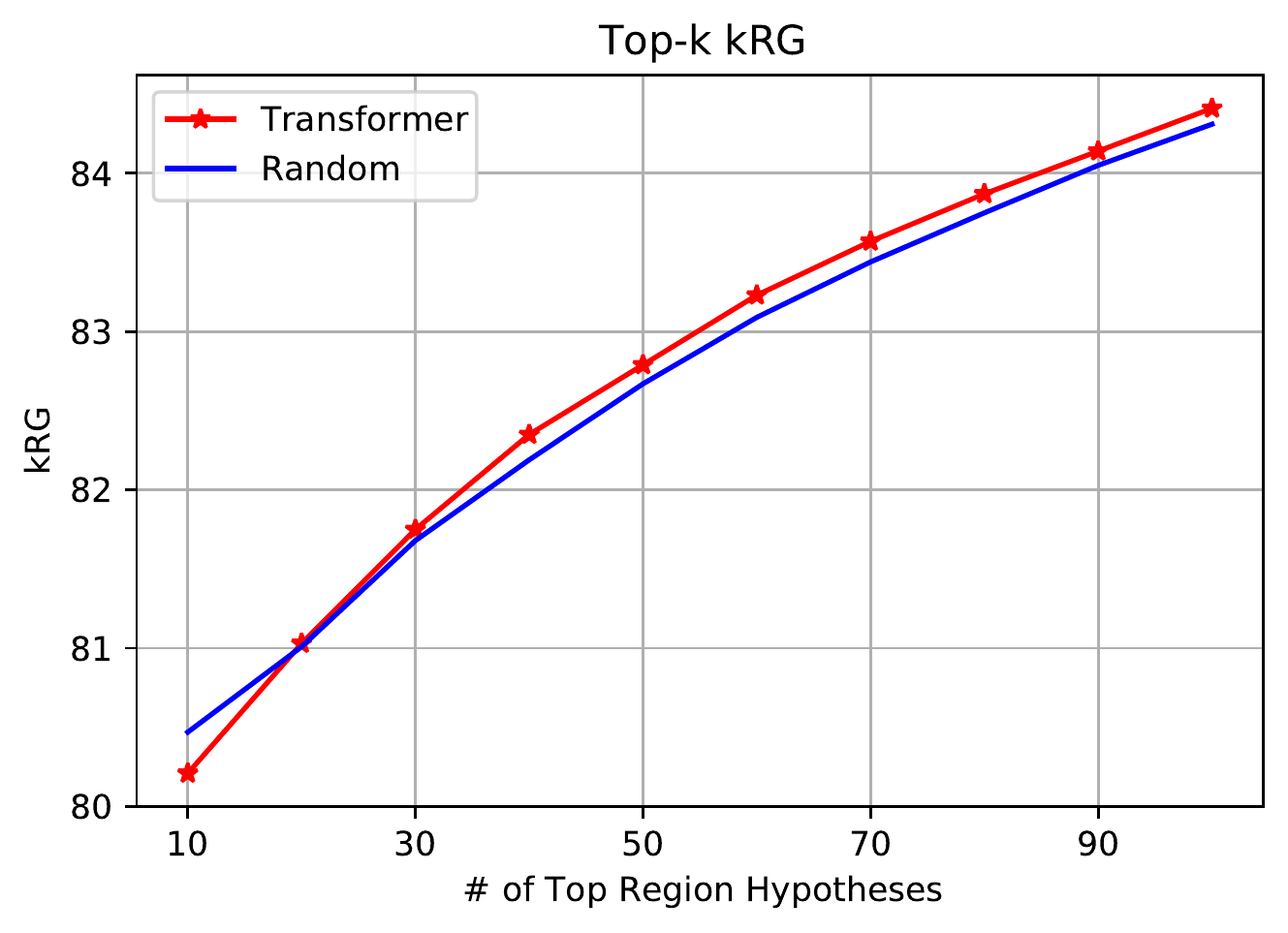}
  \vspace{-15pt}
  \label{fig:sub1}
\end{subfigure}%
\begin{subfigure}{.5\linewidth}
  \centering
  \includegraphics[width=\linewidth]{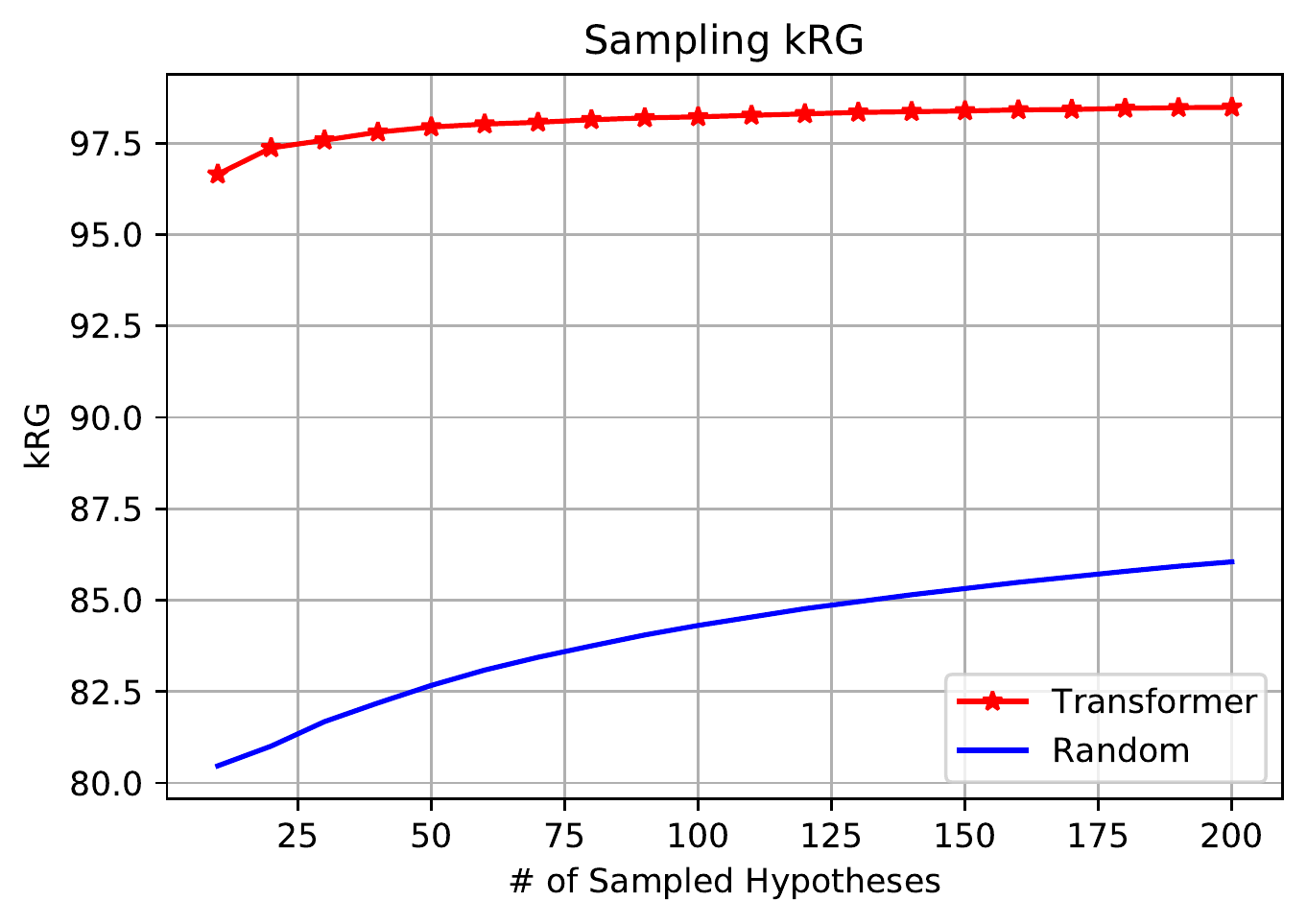}
    \vspace{-15pt}
  \label{fig:sub2}
\end{subfigure}
\caption{Ranking ability with respect to increasing number of $k$. \textbf{Left}: Top Region; \textbf{Right}: Sampling.}
\label{fig:krg_vs_k}
\vspace{-10pt}
\end{figure}

{
\vspace{3pt}\noindent\textbf{\emph{{2}. The State-of-the-art Transformer models face serious ranking problem in top region.}}
In Figure \ref{fig:krg_vs_k}, we plot the kRG results for top region and sampling. To further investigate the results, we also plot a random $\text{kRG}$. Recall definitions in Equation (\ref{eq:rank_nDCG}). The list of relevance scores $f(y_j)$ is a certain permutation of $[0, 1, \cdots, k-1]$. The random results are averaged from 100k samples of permutations. 

For top region model errors shown on the left, the model's kRG values are close to the random line when increasing $k$.
Such behavior indicates severe ranking errors, and the model performs only at the random chance level in the top region. 
In contrast, by studying the sampled results on the right, the model outperforms the random line with a considerable gap. 
The model's opposite behaviors from the top region and sampling approximation are surprising. 
We conjecture that the NMT model can distinguish good/bad hypotheses coarsely but fail at the top region and fine-grained levels. 

The above findings provide another explanation on why MBR decoding \cite{eikema2020map, freitag2021minimum} achieved better performance recently, as the model can better rank the sampling outputs. 
Rank-sensitive training \cite{chiang2012hope} might be a possible solution for the ranking errors.
}

\begin{figure}
\centering
\begin{subfigure}{.5\linewidth}
  \centering
  \includegraphics[width=\linewidth]{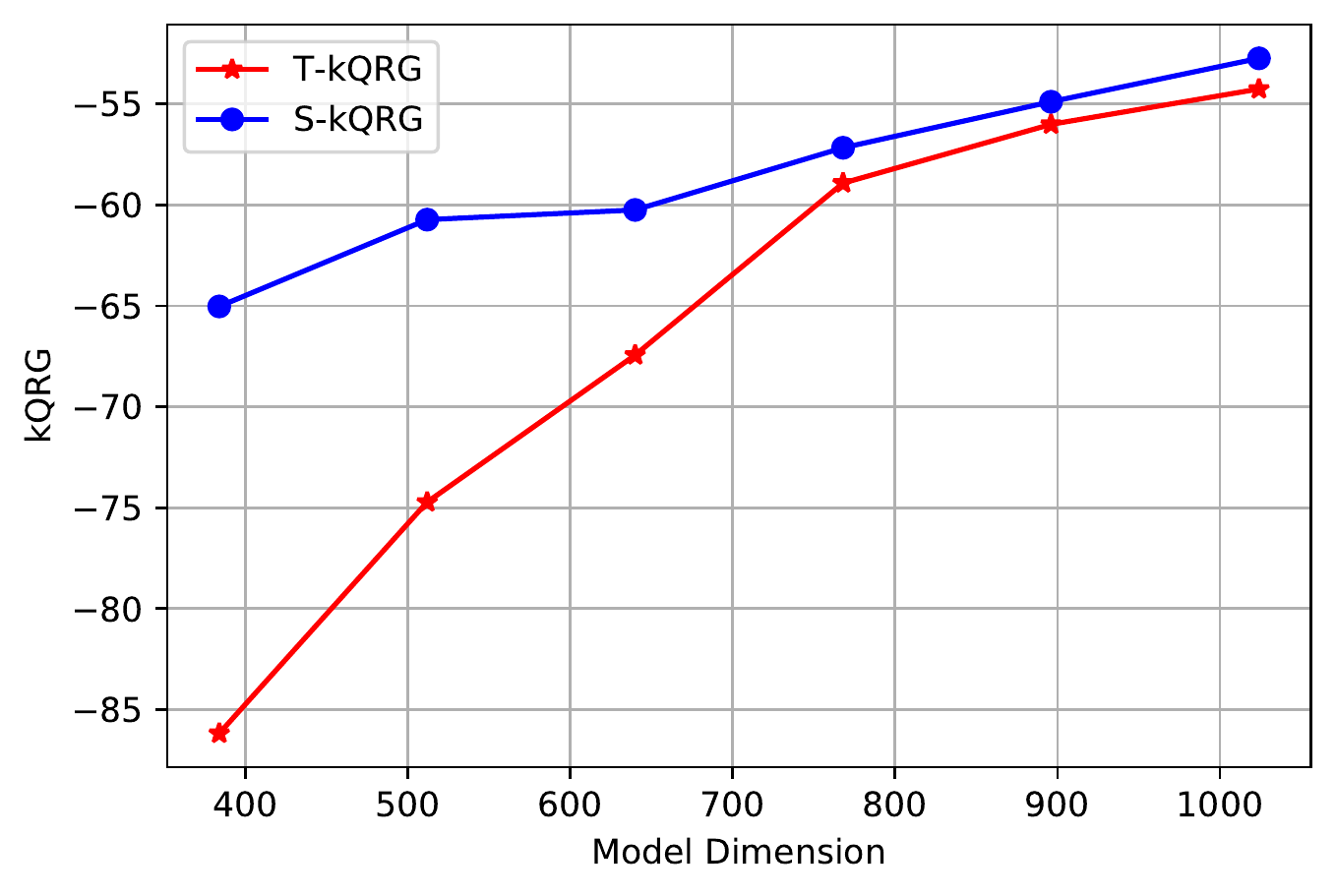}
  \vspace{-15pt}
  \label{fig:sub1}
\end{subfigure}%
\begin{subfigure}{.5\linewidth}
  \centering
  \includegraphics[width=\linewidth]{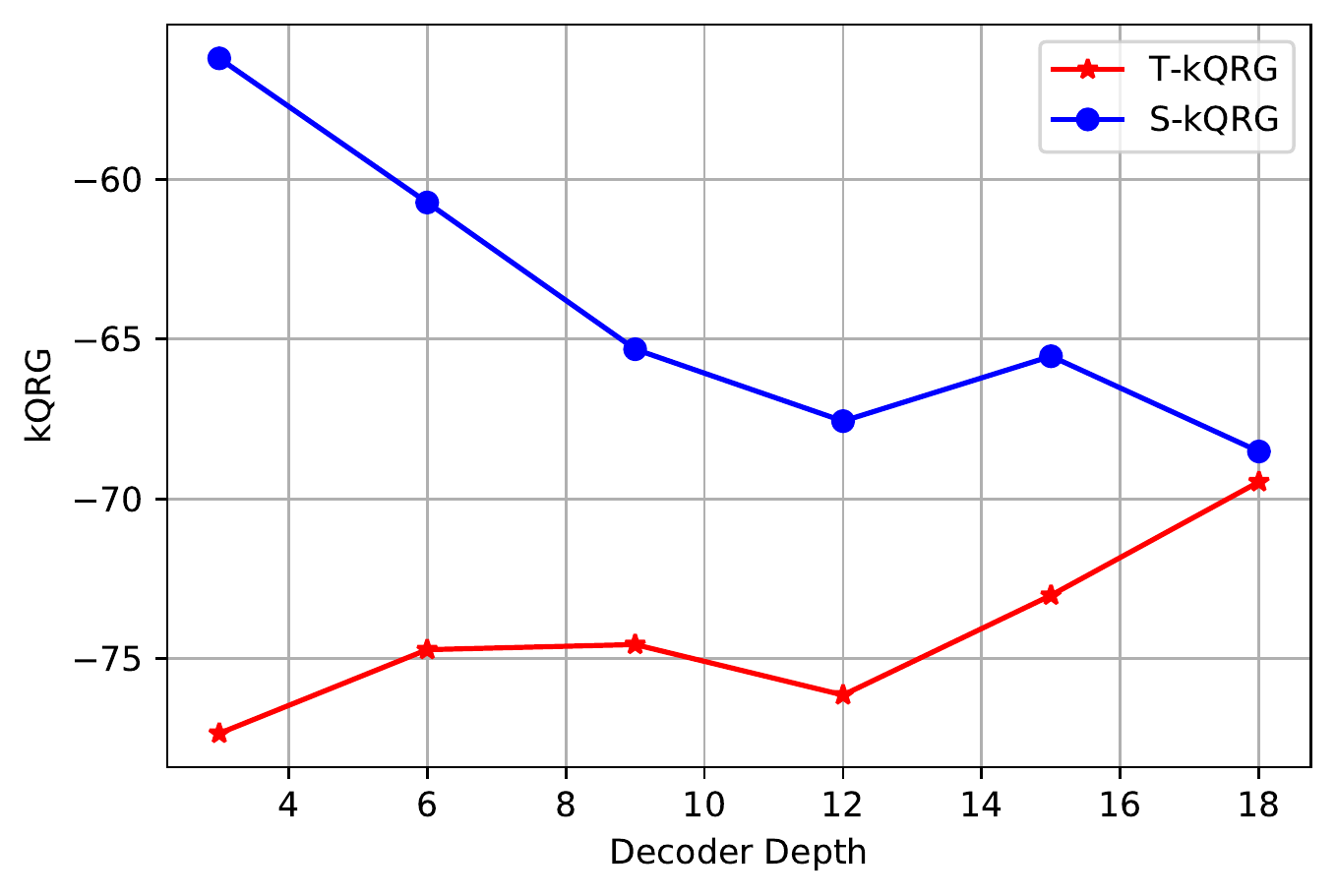}
    \vspace{-15pt}
  \label{fig:sub2}
\end{subfigure}
\caption{kQRG for Wide/Deep models. T-/S- denote Top Region/Sampling. \textbf{Left}: Model Dimension; \textbf{Right}: Decoder Depth.}
\label{fig:kQRG_wide_and_deep}
\vspace{-15pt}
\end{figure}

{

\vspace{3pt}\noindent\textbf{\emph{{3}. Widening models are more effective in reducing model errors.}} 
Recently, many interests have been drawn for using deeper models \cite{wang2019learning,li2020shallow} instead of wider models \cite{yan2020multi} to increase model capacity. Here we study the model errors of wider and deeper models. 

Our results are shown in Figure \ref{fig:kQRG_wide_and_deep}. With the increases in model dimensions, model errors with top region and sampling have both been improved. In contrast, a deeper decoder shows smaller model errors in the top region, but larger model errors in sampling (whole hypothesis space), which is counter-intuitive as we would expect that a larger model capacity means smaller model errors. 
As we do not observe a clear trend in increasing encoder depth, we put these results in the Appendix in case readers are interested. 

}



\vspace{3pt}\noindent\textbf{\emph{{4}. Model confidence may be crucial to reducing model errors.}} Results show that 
\emph{w/ para FT, w/ para BT and w/o LS} all
show impressive improvements in kQRG in both of our evaluations. 
Nonetheless, their BLEU scores with beam search are only comparable/worse than other methods like deep/wide models.
In this case, system-level evaluation fails to capture decent improvements over the model's hypothesis space.
As forward-translation training and disabling label-smoothing are expected to enhance the model confidence, we conjecture that model errors are highly related to model confidence and leave the exploration as future work.
\footnote{Due to the space limitation, we address the ablation with different base architecture, different datasets, and different origins in the Appendix. }


\begin{table}
\centering
\setlength\tabcolsep{2pt}
\begin{tabular}{l||c c | c c}
\toprule
\multirow{2}{*}{\bf Method} & \multicolumn{2}{c|}{\bf Top Region} & \multicolumn{2}{c}{\bf Beam Search} \\
\cmidrule{2-5}
& \textbf{$\text{kRG}$} & \textbf{$\text{kQRG}$} & \textbf{$\text{kRG}$} &\textbf{$\text{kQRG}$}\\
\midrule
6-layer  & 81.37 & -74.72 &	80.82\textsuperscript{-0.55} &	19.46\textsuperscript{+94.19} \\
9-layer  & 81.38 &	-74.66 &	81.16\textsuperscript{-0.22} &	20.47\textsuperscript{+95.13} \\
12-layer & 80.62 &	-66.02 &	80.84\textsuperscript{+0.22} &	22.02\textsuperscript{+88.04} \\
15-layer & 80.94 &	-66.60 &	80.90\textsuperscript{-0.04} &	22.34\textsuperscript{+88.94} \\
18-layer & 81.42 &	-73.54 &	80.69\textsuperscript{-0.74} &	22.31\textsuperscript{+95.85} \\
\midrule
\midrule
D384  & 82.05 & -86.18 & 80.59\textsuperscript{-1.46} & 16.46\textsuperscript{+102.63} \\
D512  & 81.37 & -74.72 & 80.82\textsuperscript{-0.55} & 19.46\textsuperscript{+94.19} \\
D640  & 80.82 & -67.44 & 81.16\textsuperscript{+0.34} & 21.63\textsuperscript{+89.06} \\
D768  & 80.91 & -58.92 & 81.11\textsuperscript{+0.19} & 22.28\textsuperscript{+81.21} \\
D896  & 80.20 & -56.01 & 80.14\textsuperscript{-0.06} & 22.01\textsuperscript{+78.01} \\
D1024 & 80.57 & -54.26 & 81.26\textsuperscript{+0.69} & 23.26\textsuperscript{+77.52} \\
\bottomrule
\end{tabular}
\caption{Hypothesis space evaluation over top-10 outputs versus beam top-10 outputs when increasing dimensions / enc layers .}
\label{table:wide_beam_hr}
\vspace{-15pt}
\end{table}

{

\subsection{Connection to Search Algorithms}
\subsubsection{Quantify Beam Search Lucky Biases}
As pointed out in recent work \cite{meister-etal-2020-beam}, beam search seems to bring a lucky bias that covers some of the model errors. This section utilizes our proposed metric to understand the bias brought by beam search, since our top-region evaluation finds the best solution for MAP decoding with no search errors. 

Concretely, we use kRG and kQRG to evaluate the errors from both exact top-$k$ and beam search top-$k$ outputs and compare the scores to check the effect of beam search bias. In this way, the gap between two errors represents the lucky bias brought by beam search quantitatively. Experiments are conducted in NIST Zh-En, and results are shown in Table \ref{table:wide_beam_hr}. We have several interesting findings. 

Firstly, beam search leads to a decent improvement (from +77\% to +102\%) in kQRG, which quantitatively proves the existence of beam search's lucky bias in recent work. 

Then, beam search generally does not affect ranking abilities. 
As shown, the gaps of kRG between the beam and exact outputs are generally small and fluctuate around 0. We do not observe a clear trend of beam search bias in ranking abilities. 

Furthermore, we analyze deeper and wider models and observe different behaviors. 
There is a clear trend in decreasing the gap between the beam and the exact when increasing the model's width. 
Conversely, the lucky biases of beam search retain when increasing the model's depth, showing deeper models are more compatible with beam search biases than wider ones.
Such behaviors concur with the studies, showing that deeper models perform more efficiently and effectively with beam search than wider models \cite{wang2019learning,li2020shallow}.
We show that the observed superiority of deep models may stem from their compatibility with beam search's inductive bias. 


\begin{table}[]
\centering
\begin{tabular}{l||c|c}
\toprule
\textbf{Method}  & \textbf{Pearson} & \textbf{Cost}         \\ 
\hline
MBR         & 1.000       & $N^2$ \\ \hline \hline
Beam Search & -0.143          & $N$                    \\ 
Sampling    &   0.975              & $N$                    \\ 
Ours        &  \bf{0.977}        & $N$                    \\ 
\bottomrule
\end{tabular}
\caption{Correlation studies for MBR decoding.}
\label{tab:mbr_corr}
\vspace{-15pt}
\end{table}

\subsubsection{Correlations with MBR Decoding}
MBR decoding emerges as a promising and powerful decoding algorithm instead of beam search \cite{eikema2020map,freitag2021minimum}, which makes use of sampled hypothesis space and is relevant to our proposed evaluation. Here, it is necessary to study the correlation between our proposed sampling evaluation and MBR decoding.

Concretely, we perform experiments over our ten systems with WMT'14 En-De, and we test the Spearsman/Kendall correlation between MBR decoding translation qualities and our sampled kQRG scores. For MBR, we use 100 samples per source sentences and BLEURT \cite{sellam2020bleurt} as our utility function, following \citet{freitag2021minimum}.
One salient advantage of our proposed evaluation instead of directly MBR over test sets is the computational cost. 
For instance, with 100 samples, our evaluation uses 100 BLEURT calls per sentence, while the naive MBR needs 10k BLEURT calls due to its usage of quadratic computations. 

We also report the correlation for the other two evaluations, namely beam search and sampling. As shown in Table \ref{tab:mbr_corr}, our method performs the best among the three evaluations, and it indicates a potential application for our sampling-based kQRG.

}

\color{black}
\section{Related Work}
\textbf{Decoding Methods.}
Most decoding methods in NMT aims to find the hypothesis with the highest conditional probability, i.e., maximum-a-posterior~(MAP) decoding.
Among all MAP decoding methods, beam search is most widely applied in the modern NMT systems for evaluation. 
Naive beam search has several known drawbacks, such as favoring short translations and its monotonic constraint. 
Hence, many regularization/rescoring methods~\citep{bahdanau2014neural,wu2016google,he2016improved,yang2018breaking,murray2018correcting} or beam search variants~\citep{freitag2017beam,shu2018improving} are proposed to improve the performance.
Other than beam search, one promising MAP decoding for evaluation is the DFS-based exact search \citep{stahlberg-byrne-2019-nmt}, which finds the mode of model distributions. Despite its high computational cost, it reveals important information about the learned hypothesis space.  
We follow this approach and present a top-$k$ exact search method, which can access the top-region of hypothesis space. 

In addition, there are some non-MAP decoding algorithms. 
A typical one is the stochastic sampling-based decoding methods \citep{ackley1985learning,holtzman2019curious}, which randomly choose candidates from each step's output distribution.
Further, \citet{eikema2020map} introduces a Minimum Bayesian Risk decoding method based on sampling. \citet{leblond2021machine} propose a metric-driven search approach via Monte-Carlo Tree Search (MCTS). The sampling-based methods are promising and may incorporate with our evaluation in future directions.

\noindent\textbf{Error Evaluation.}
Evaluation of NMT errors focuses on studying the gap between machine-translated results and human-translated references. 
Statistical matching metrics \citep{papineni2002bleu, banerjee2005meteor, koehn2007moses,denkowski2014meteor,guo2019meteor++} and pretrained metrics \cite{sellam2020bleurt, rei2020comet} are two dominant directions in evaluating errors. These metrics prove that linguistic similarity between references and machine translations correlates the human evaluation well. 
However, to the best of our knowledge, these statistical metrics evaluate one best hypothesis decoded from heuristic decoding algorithm (i.e., system-level evaluation), which incorporate huge search errors and bias understanding of NMT models. 

Recent efforts \citep{niehues-etal-2017-analyzing, stahlberg-etal-2018-versatile, stahlberg-byrne-2019-nmt, meister-etal-2020-beam,eikema2020map} are devoted to analyzing model errors without search errors and provide meaningful conclusions.
Nonetheless, these approaches still evaluate over one hypothesis in hypothesis space except with the one with highest probability.
This is incomprehensive due to neglecting errors inside the whole hypothesis space.
In contrast, we dig into model errors over top regions and provide a more comprehensive evaluation. In addition, we provide various interesting findings over model errors with regards to NMT techniques and search algorithms.

\section{Conclusion}
{
This paper presents a novel evaluation protocol for model errors in the perspective of rankings over the hypothesis space. 
Specifically, our evaluation encompasses two approximated evaluations, top region and Monte Carlo Sampling, and two metrics, kRG and kQRG, measuring the hypothesis ranking ability of hypothesis space.
Our evaluations correlate well with human judgments and provide interesting findings over NMT techniques and search algorithms. 
We believe these findings shed light on future development in the NMT field. 

For future directions, we think the evaluation of NMT models should disentangle with search algorithms, and assess models more comprehensively from the perspective of hypothesis space. Furthermore, the effectiveness of different NMT techniques should also be re-evaluated from such a perspective. 
We expect multi-angle evaluations over the NMT models. 
Errors we revealed, like the ranking errors, need to be fixed and may have connections with the well-known beam search curse problem, which is also a promising direction worth exploring.}

{
}





\bibliography{neurips_2021}
\bibliographystyle{acl_natbib}

\clearpage
\appendix
\clearpage
\appendix
\addcontentsline{toc}{section}{Appendix} 
\part{Appendix} 
\parttoc 

\section{Correlation with Human Judgements}
{
\label{sec:human_corr}
This section provides the correlation results for different choices of translation quality metrics. We choose four reference-based metrics: sentence-BLEU, ChrF, BLEURT, and COMET, and a reference-free QE metric: COMET-QE. We test both the sentence and system score correlations between kQRG and human judgments. The results are shown in Table \ref{table:human_corr_all}. 

Among all translation quality metrics, sentence-BLEU performs the worst, and COMET shows the strongest correlation in both sentence and system levels. This justifies our choice of COMET for the main results. We also find that ChrF has good correlations with human evaluation. Therefore, we provide results for ChrF in the following sections. In addition, our evaluation can be incorporated with QE metrics and becomes a reference-free evaluation protocol. However, COMET-QE lags behind other reference-based translation quality metrics in terms of correlation.}
\begin{table*}[]
\centering
\begin{tabular}{@{}l|ll|ll@{}}
\toprule
\multicolumn{1}{c|}{\multirow{2}{*}{Translation Quality}} & \multicolumn{2}{c|}{Sentence}          & \multicolumn{2}{c}{System}             \\ \cmidrule(l){2-5} 
\multicolumn{1}{c|}{}                                     & \multicolumn{1}{l|}{Pearson} & Spearman & \multicolumn{1}{l|}{Pearson} & Spearman \\ \midrule
Sentence-BLEU                                             & \multicolumn{1}{l|}{0.67}   & 0.80     & \multicolumn{1}{l|}{0.59}   & 0.55     \\ \midrule
ChrF                                                      & \multicolumn{1}{l|}{\underline{0.85}}   & \bf 0.86     & \multicolumn{1}{l|}{\underline{0.75}}   & \underline{0.72}     \\ \midrule
BLEURT                                                    & \multicolumn{1}{l|}{\bf 0.86}       &  \underline{0.85}   & \multicolumn{1}{l|}{0.71}       &  0.61   \\ \midrule
COMET                                                     & \multicolumn{1}{l|}{\bf 0.86}   & \underline{0.85}     & \multicolumn{1}{l|}{\bf 0.78}   & \bf 0.82     \\ \midrule
COMET-QE                                                  & \multicolumn{1}{l|}{0.66}   & 0.66     & \multicolumn{1}{l|}{0.71}   & 0.53     \\ \bottomrule
\end{tabular}
\caption{Pearson and Spearman's correlation scores with human judgements across different translation quality functions. Bold and underline represent the 1st and 2nd performing results, respectively. }
\label{table:human_corr_all}
\end{table*}

\begin{table}
\centering
\setlength\tabcolsep{2pt}
\begin{tabular}{l|r|c|c|c}
\toprule
\bf Name & \bf Train & \bf Dev & \bf Test & \bf BPE \\
\midrule
NIST Zh-En & 1.2M & 1664 & 5105 & 40K/30K \\
\midrule
WMT'14 En-De & 4.5M & 3000 & 3003 & 32K \\
\midrule
WMT'14 En-Fr & 35.7M & 6003 & 3003 & 40K \\
\bottomrule
\end{tabular}
\captionof{table}{Statistics of Datasets}
\vspace{-10pt}
\label{table:stats_of_datasets}
\end{table}

\section{Experimental Details}
\label{sec:exp_details}
\subsection{Detailed Descriptions of Datasets}
For our WMT'14 En-De/En-Fr tasks, we use 4.5M / 35.7M preprocessed data, which is tokenized and split using byte pair encoded (BPE) \citep{sennrich-etal-2016-neural} with 32K/40K merge operations and a shared vocabulary for source and target sides. For En-De, we use \emph{newstest2013} as the validation set and \emph{newstest2014} as the test set. For En-Fr, we use the combination of \emph{newstest2012} and \emph{newstest2013} as our validation set and \emph{newstest2014} as the test set.

For the NIST Zh-En task, we use 1.25M sentences extracted from LDC corpora\footnote{The corpora include LDC2002E18, LDC2003E07, LDC2003E14, Hansards portion of LDC2004T07, LDC2004T08 and
LDC2005T06.}. 
To validate the performance of our model, we use the NIST 2006 (MT06) test set with 1664 sentences as our validation set.
Then, the NIST 2002 (MT02), 2004 (MT04), 2005 (MT05), 2008 (MT08) test sets are used as our test sets, which contain 878, 1788, 1082, and 1357 sentences, respectively. All reported results are averaged over different test sets.

The statistics of all three datasets can be found in Table \ref{table:stats_of_datasets}.

\subsection{Training Details}
Our models are trained using the \emph{fairseq} toolkit\footnote{\url{https://github.com/pytorch/fairseq}}. We train each of our Transformer models for 100k/300k/300k steps for three datasets and validate every 5000 steps. 
The default label smoothing is $0.1$. The dropout rates for different Transformer models range from $0.1$ to $0.4$. 
The batch sizes are 8k/64k/64k tokens for three datasets.
All our Transformer models are pre-norm models. 
Other hyperparameter settings are the same as in \citep{vaswani2017attention}.
For evaluation, we report case-sensitive tokenized BLEU scores using \emph{multi-bleu.perl}\footnote{\url{https://github.com/moses-smt/mosesdecoder/blob/master/scripts/generic/multi-bleu.perl}} for both WMT'14 En-De and En-Fr, and case-insensitive tokenized BLEU scores for NIST Chinese-English. 
We select the best checkpoint on the validation set and report its performance on the test set. All reported results are averaged over all sentences in the test set. For results with beam search, the beam size is 5, and the length penalty is 0.6.

\section{Additional Experimental Results on Model Errors}
\label{sec:additional_results}
\subsection{Various NMT Benchmarks}
\label{sec:comet_across_lang}
{

This section presents COMET results on the WMT'14 English-French and NIST Chinese-English tasks. 
The results are shown in Table \ref{table:zhen}, \ref{table:enfr}. It is encouraging that the results are all consistent and corroborate our findings in the main text. As these three datasets have small, medium, and large sizes, we prove that our proposed protocol generalizes well across different languages and sized datasets. 

Furthermore, by comparing results among these experiments, we find that model errors for different tasks vary vastly. The reason might be either the intrinsic difficulties of tasks or other properties of the dataset like sizes or cleanliness, etc. We revisit the dataset properties in Section \ref{sec:dataset_prop}. 
}
\begin{table}[t!]
\setlength\tabcolsep{2pt}
\begin{tabular}{l||c ||c | c c}
\toprule
\multirow{2}{*}{\bf Method} & \bf System &\textbf{{Mode}} & \bf Top & \bf Sample\\
\cmidrule{2-5}
& \textbf{BLEU} & \textbf{\# Emp} & \textbf{$\text{kQRG}$} & \textbf{$\text{kQRG}$} \\
\midrule
Transformer & 42.47 & 41.14 & -74.72 & -60.73 \\
\midrule
w/o LS & 42.44 & \bf 14.59 & -31.78 & \bf{-11.54} \\
w/ para FT & 42.17 & 17.52 & \bf{-23.42} & -52.83\\
\midrule
w/ 12-layer Enc & 43.38 &  36.24 & -66.02 & -59.63 \\
w/ 18-layer Enc & 43.81 &  43.11 & -73.54 & -58.85 \\
\midrule
w/ Dim 768 & 42.88 & 40.76 &  -58.92 & -57.17 \\
w/ Dim 1024 & \bf{43.43} & 34.03 & -54.26 & -52.74 \\
\bottomrule
\end{tabular}
\caption{COMET model errors of different models in NIST Chinese-English task. kQRG values are not normalized. }
\label{table:zhen}
\end{table}

\begin{table}[tb!]
\setlength\tabcolsep{2pt}
\begin{tabular}{l||c ||c | c c}
\toprule
\multirow{2}{*}{\bf Method} & \bf System &\textbf{{Mode}} & \bf Top & \bf Sample\\
\cmidrule{2-5}
& \textbf{BLEU} & \textbf{\# Emp} & \textbf{$\text{kQRG}$} & \textbf{$\text{kQRG}$} \\
\midrule
Transformer & 40.78 & 46.75 & -22.69 & -96.48 \\
\midrule
w/o LS & 40.70 & \bf 19.51 & 28.37 & \bf{~5.69} \\
w/ para FT & 40.95 & 27.26 & \bf{49.67} & -82.75 \\
\midrule
w/ 12-layer Enc & 41.28 & 44.99 & -18.96 &	-95.62 \\
w/ 18-layer Enc & 41.74 & 53.58 & -16.91 &	-94.56 \\
\midrule
w/ Dim 768 & 41.73 & 46.12 & -17.71 &	-93.17 \\
w/ Dim 1024 & \bf{42.35} & 40.42 & -11.04 &	-87.07 \\
\bottomrule
\end{tabular}
\caption{COMET model errors of different models in WMT'14 En-Fr task. kQRG values are not normalized. }
\label{table:enfr}
\end{table}

\subsection{Various Translation Quality Functions}
\label{sec:various_quality_function}
{

This section provides model errors with an additional reference-based translation quality metric -- ChrF, which performs second to COMET in our correlation studies. 

In Table \ref{table:ende_ChrF}, we present our results using ChrF with the English-German task. 
An advantage of using ChrF is its bound between 0 and 1, which makes our kQRG easier to interpret. 
We observe that all of our findings in Section \ref{sec:find_techinique} still hold. 
This proves our proposed protocol performs consistently across different choices of translation metrics.

}

\begin{table}
\centering
\setlength\tabcolsep{2pt}
\begin{tabular}{l||c || c | c c}
\toprule
\multirow{2}{*}{\bf Method} & \bf System &\textbf{{Mode}} & \bf Top & \bf Sample\\
\cmidrule{2-5}
& \textbf{BLEU} & \textbf{\# Emp} & \textbf{$\text{kQRG}$} & \textbf{$\text{kQRG}$} \\
\midrule
Transformer & 27.22 & 64.70 &   31.67 & 34.58 \\
\midrule
w/o LS & 26.76 & 34.85 &        41.64 & 42.41 \\
w/ para BT & 27.36 & 27.26 &    42.89 & 43.31 \\
w/ para FT & 28.06 & \bf{0.93} &\bf{55.55} & \bf{49.72} \\
\midrule
w/ 12-layer Enc & 27.75 & 58.11& 33.86 & 35.11 \\
w/ 18-layer Enc & 28.03 & 53.58& 35.33 & 36.48  \\
\midrule
w/ Dim 768 & 28.00 & 50.18 &35.60 & 35.67  \\
w/ Dim 1024 & \bf{28.49} &44.72& 37.75 & 37.99  \\
\bottomrule
\end{tabular}
\caption{ChrF model errors of different models in WMT'14 English-German.}
\label{table:ende_ChrF}
\end{table}


\subsection{Various Model Architectures}
{
In previous sections, we discuss the model errors of Transformer models. In this section, we extend the experiments to different NMT architectures, i.e., ConvSeq2Seq \cite{gehring2017convolutional} and RNNSearch \cite{luong2015effective}. We use the WMT'14 English-German and present our model error (COMET) results in Table \ref{table:other_arch}.

Interestingly, we find that RNNSearch performs the best in terms of kRG, indicating the strongest ranking capability. ConvSeq2Seq has a 63.08 score in kRG and is second to RNNSearch. Both of them perform better than the Transformer model in terms of ranking capability and are better than random ranking (58.72 in Section \ref{sec:find_techinique}).
Then, the Transformer model outperforms ConvSeq2Seq and RNNSearch in terms of model error and BLEU score, showing a stronger hypothesis selection. 
On the one side, these results demonstrate that future model design needs to revisit RNN models' advantages and incorporate them with current Transformer architectures. On the other side, the RNN model with the best ranking ability only scores $66.16$ of $[0, 100]$ in kRG, indicating large potentials in reducing model errors by improving their ranking abilities. 
}
\begin{table}[tb!]
\centering
\setlength\tabcolsep{2pt}
\begin{tabular}{l||c ||c | c c}
\toprule
\textbf{Method} & \textbf{BLEU} &  \textbf{$\text{kRG}$} & \textbf{$\text{kQRG}$} \\
\midrule
Transformer & \bf{27.22} &80.21 & \bf{-60.39} \\
\midrule
RNNSearch &23.07 & \bf{83.63} & -106.26 \\
ConvS2S &26.51 & 81.76 & -77.40 \\
\bottomrule
\end{tabular}
\caption{Top region model errors with different model architectures in WMT'14 English-French.}
\label{table:other_arch}
\end{table}

\subsection{Analysis on Original Sources}
{
One interesting result in our main experiments is that the \emph{paraFT} model performs much better than the \emph{paraBT} model. One possible reason is that \emph{paraFT} model overfits the original sides of the test sets. 
Therefore, we compare model errors on the English-original part and German-original part of \emph{newstest2014} to verify this assumption, which contains 1,500 and 1,503 sentences, respectively. 

Table~\ref{table:orig_lang} shows the results. Comparing "Source En" with "Source De", we find that the ranking capabilities (kRG) are not much affected by the original sides. However, models perform substantially better in kQRG of source German side than that of the source English side, as translated English sentences are easier to translate than original English sentences. 
The gap between \emph{paraFT} and \emph{paraBT} varies to some extent across different origins, but with both sides, \emph{paraFT} still strongly outperforms \emph{paraBT}. Thus we conclude that original side is not the main reason. 
}
\begin{table}
\centering
\setlength\tabcolsep{2pt}
\begin{tabular}{l||c | c | c | c}
\toprule
\multirow{2}{*}{\bf Method} & \multicolumn{2}{c|}{\bf Source En} & \multicolumn{2}{c}{\bf Source De} \\
\cmidrule{2-5}
& \textbf{$\text{kRG}$} & \textbf{$\text{kQRG}$} & \textbf{$\text{kRG}$} &\textbf{$\text{kQRG}$}\\
\midrule
Transformer base       &  80.84 & -84.46 & 79.58 & -36.24 \\
 w/ para ft &  81.87 & 35.98 & 82.42 & 50.61 \\
 w/ para bt &  79.47 & -33.76 & 80.46 & ~7.50 \\
Transformer Big	       &  80.63 & -59.59 & 80.59 & -9.46 \\
\bottomrule
\end{tabular}
\caption{Top region model errors on English-original and German-original part of newstest2014 En-De testset.}
\label{table:orig_lang}
\end{table}

\subsection{Clean and Up-to-date Datasets}
{
\label{sec:dataset_prop}
There is a concern that the ranking issues are from the WMT'14 datasets, which are outdated and noisy \cite{pmlr-v80-ott18a}. In this section, we study properties of the datasets and provide two additional ablation experiments to support our method. We introduce two datasets: (1) WMT'14 En-De dataset filtered by language detection and the fast align, (2) the WMT'20 En-De dataset, to which we perform the same filters.
These two datasets contain 3.86M and 37.2M paired sentences, respectively. 
For language detection, we use the pre-trained fasttext tool \footnote{\url{https://github.com/facebookresearch/fastText}} and filter out the sample if either side of a paired sentence is identified as other languages. 
For the fast align \cite{dyer2013simple} filtering, we compute both the source-target and target-source alignment scores and filter out sentences with an average score less than $-6$. 

The results are shown in Table \ref{table:ende_new_datasets}. We have four key observations. 
Firstly, by comparing original WMT'14 En-De results with datasets after language detection (LD) and fast align filtering (FA), we find fine-grained cleaning techniques help reduce model errors. The kQRG values improve significantly, from -60.39\% to -45.41\%. 
Secondly, training with an up-to-date dataset dramatically improves the model in terms of reducing errors. As the WMT'20 En-De training set (37.2M) is much larger than the WMT'14 En-De (4.5M), we also conduct experiments with different sampled sizes of the WMT'20 dataset from 4.5M to 20M. We find that even with the same training set size (4.5M), the model trained with the WMT'20 dataset outperforms its WMT'14 counterpart (-21.02\% versus -29.95\%). 
Thirdly, we attempt to increase the size of training corpus with WMT20 En-De. Surprisingly, we observe that top region model errors go slightly up. 
Fourthly, all our models with clean or updated datasets still do not show stronger ranking abilities than random rankings. 

All above findings reveal two points: (1) The ranking errors we identified in the main text still exist even with cleaner or up-to-date datasets. The main cause for these ranking problems is not the training set. (2) Using a clean, up-to-date dataset reduces model errors. It helps the model move better hypotheses into the top-region of hypothesis space, thus achieving better kQRG scores. The results for kQRG values are strongly dependent on the datasets.
}


\begin{table}
\centering
\setlength\tabcolsep{2pt}
\begin{tabular}{l||c | c}
\toprule
\bf Dataset & \textbf{$\text{kRG}$} & \textbf{$\text{kQRG}$} \\
\hline
WMT'14 En-De (4.5M) & 80.21 &-60.39 \\
\midrule
w/ LD & 80.24 & -50.88 \\
w/ LD + FA & 80.32 & -45.41 \\
\midrule \midrule
WMT'20 En-De (37M) & 80.19 & -21.84 \\
\midrule
w/ Sample 4.5M & 79.88 & -21.02 \\
w/ Sample 10M & 80.09 & -23.42 \\
w/ Sample 20M & 80.39 & -29.95 \\
\bottomrule
\end{tabular}
\caption{Top region model errors over filtered WMT'14 En-De and WMT'20 En-De tasks. The model we use is the Transformer-base model. LD denotes filtering with language detection. FA denotes filtering with fast align.}
\label{table:ende_new_datasets}
\end{table}

\subsection{Increasing encoder depth}
\begin{figure}[t]
    \includegraphics[width=0.9\linewidth]{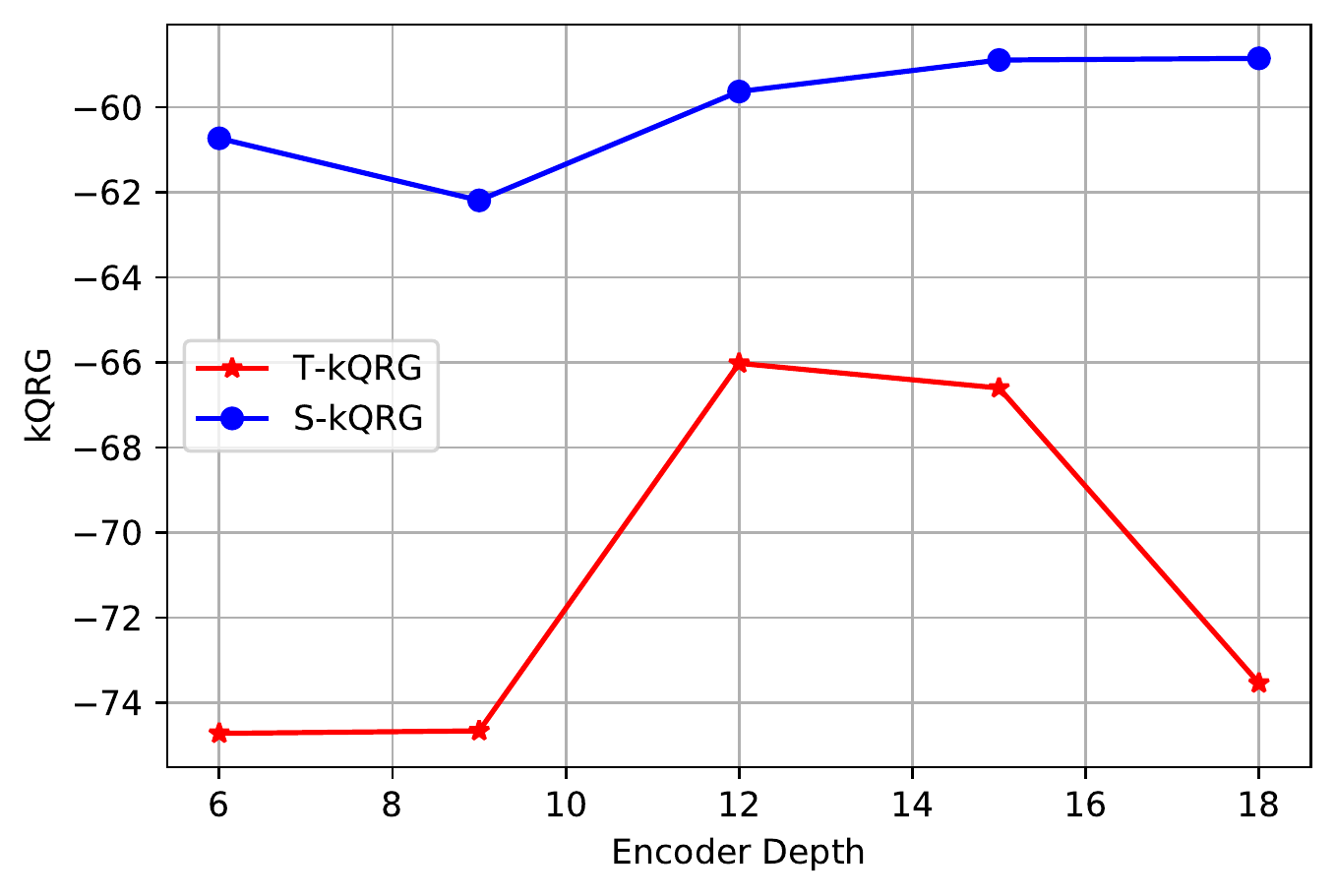}
\caption{kQRG for deep encoder models.}
\label{fig:deep_enc}
\end{figure}

As discussed in Section \ref{sec:find_techinique}, we plot the model errors for deep encoders in Figure \ref{fig:deep_enc}. We do not observe a clear trend for smaller or larger model errors when increasing encoder depth. 

\section{Implementation Details of Exact Top-$k$}
\label{sec:computation}
Here we explain the implementation details of our exact top-$k$ algorithm. The detailed algorithm is shown in Algorithm \ref{dfs_topk_appendix}. 
Our implementation is built upon \emph{uid-decoding}\footnote{\url{https://github.com/rycolab/uid-decoding}} and \emph{sgnmt\footnote{\url{https://github.com/ucam-smt/sgnmt}}} projects, and is compatible with the models trained with \emph{fairseq}.
The original implementation of exact top-$1$ decoding heavily relies on CPU operations. In contrast, our top-$k$ version moves a number of computations to GPU, and improves several implementation details as follows.
\begin{itemize}
    \item \textbf{Optimizing the iterating process.} As defined the $13$-th line of our Algorithm \ref{dfs_topk_appendix}, we need to iterate through all words in the vocabulary. However, the order of iterations significantly influences the speed because of the lower bounds. Empirically, we find that iterating the vocabulary greedily substantially reduces the run time. 
    \item \textbf{Batching the hypotheses for each time step.} As stated at the $14$-th line of Algorithm \ref{dfs_topk_appendix}, we iterate one word and perform one forward model inference at a time. However, the GPU utilization of this scheme is extremely low. Thus, we use batch technique and batch $\mathbf{b}$ different words for one model forward pass, which efficiently increases the GPU utilization.
    \item \textbf{Good lower bounds facilitate the search process.} We observe that better lower bounds vastly reduce the search time. In our implementation, we use the top $n$-best list output from the beam search with larger beam sizes than $n$ as our lower bounds.
\end{itemize}
\noindent As a result, the speed is improved significantly.

\subsection{Worst-case Analysis for Exact Search Algorithm}
This section analyzes the worst-case behaviors of exact search algorithms.
First, let us discuss a simple case when the exact search does not use lower bounds. 
Given a target sentence set $Y_l=\{ y | \text{len}(y)=l \}$ where all hypotheses in that set have the same length $l$, it is obvious that the search operations needed for exact top-$1$ and exact top-$k$ algorithms are the same, i.e., $N_l=|Y_l| = |V|^l$. Thus, the total search operations for all lengths\footnote{We do not use the length constraint in our implementation. Here, we add the max length constraint for clarity.} $l \in [1, l_{\text{max}}]$  can be computed by $N=\sum_{l \in [1, l_{\text{max}}]} N_l$. 

Next, we consider the case with lower bounds. 
Since lower bounds help trim the search space, the worst case happens when the search algorithm finds the hypotheses in a reversed order. In that case, lower bounds could not trim any search space and have to iterate all hypotheses. Hence, the numbers of search operations needed for both top-$1$ and top-$k$ algorithms are identical, i.e., $N=\sum_{l \in [1, l_{\text{max}}]} N_l$ operations.
On the other hand, both the top-$1$ and our top-$k$ algorithms are similar to Branch\&Bound algorithm \citep{hendy1982branch}, which cannot lower the time complexity in the worst case, and its time complexity is the same as the one of depth-first-search (DFS) algorithm \citep{Alan13}. However, it is practically useful because it is proved to be able to improve the search speed significantly.

\begin{algorithm}[t]

\LinesNumbered
\SetAlgoLined
\SetKwInOut{Input}{Input}
\SetKwInOut{Output}{Output}
\Input{x: Source sentence, y: Translation prefix (default: []),
p: $\log P(y|x)$ (default 0.0), k: Top-k hypotheses to output}
\Output{List $l$ contains top-k hypotheses with log-probabilities.}

\SetKwFunction{FDFS}{dfsTopK}

\textbf{global} minHeap

\textbf{global $\gamma \gets -\inf$}
  
\SetKwProg{Fn}{Function}{:}{}
\Fn{\FDFS{$x$, $y$, $p$}}{
\If{$y[|y|-1] = </s> $}
{
    push(minHeap, $(p, y)$)
    
    \If{$\text{len}(\rm{minHeap}) > k$}{ pop(minHeap)}
    
    \If{$\text{len}(\rm{minHeap}) = k$}{$\gamma \gets \rm{minHeap[0][0]}$}
    
}

\For{$v \in V$}{
    
    $p' \gets p + \log P(v|x, y)$
    
    \If{$p' \geq \gamma$}{
        \FDFS{$x, [y;v], p' $} 

    }
}
\Return minHeap
}

\Return \FDFS{$x, [], 0.0 $} 
\caption{DFS-based Top-k Exact Search.}
\label{dfs_topk_appendix}
\end{algorithm}




  
    
    
    





    





\subsection{Empirical Computational Cost}
This section provides several empirical results to show how different decoding methods perform in terms of computational time. We randomly sample 100 sentences in WMT'14 En-De \emph{newstest2014} and report the corresponding run time as well as the number of expansion operations. The expansion operation, i.e., model's forward pass, is the most time-consuming operation in the exact search algorithm and is linear to the number of computation flops.
We report the computational costs for three different algorithms, including \emph{Beam Search}, \emph{Exact Top-1} and \emph{Exact Top-5}. Each reported number is the average over four runs with different samples as inputs. 
    
The results are shown in Table \ref{table:time_cost}. First, we can see that \emph{Beam Search} is about ten to twenty times faster than exact search algorithms. This is consistent with results in previous literature. Second, compared with previous Exact Search implementation, our implementation of top-$5$ search has almost the same time cost as top-$1$, which demonstrates the effectiveness and efficiency of our proposed approach.

By taking the number of expansions into account, we notice two more interesting facts -- On the one hand, the number of expansions is not linear to $k$. Our top-$k$ algorithm explores only about five times the search space compared with top-$1$ algorithm. On the other hand, our algorithm is significantly more efficient than the original implementation, with four times faster in terms of the number of expansions and only about two times in terms of the computational cost. { In our own experiments, we use 8 NVIDIA V100 GPUs for decoding, and it takes about a day to decode exact top-10 on a standard WMT testset.}

\begin{table*}
\centering
\setlength\tabcolsep{2pt}
\begin{tabular}{l||c | c}
\toprule
\bf Method & \bf Time Cost (seconds) & \bf Num Expansions\\
\midrule
Beam Search & 453.0 & - \\
\citet{stahlberg-byrne-2019-nmt} & 8,064.0 & 2,769.6 \\
\midrule
Exact Top-$5$ w/ BS lower bounds & 8,914.4 & 6,029.4 \\
\bottomrule
\end{tabular}
\caption{Time cost and number of expansions for exact search algorithms with 4 sampled runs on 100 test sentences.}
\label{table:time_cost}
\end{table*}

\subsection{Choice of Different $k$ Values}
\label{sec:different_k}

\begin{figure}[t]

\subfloat{%
    \includegraphics[width=0.9\linewidth]{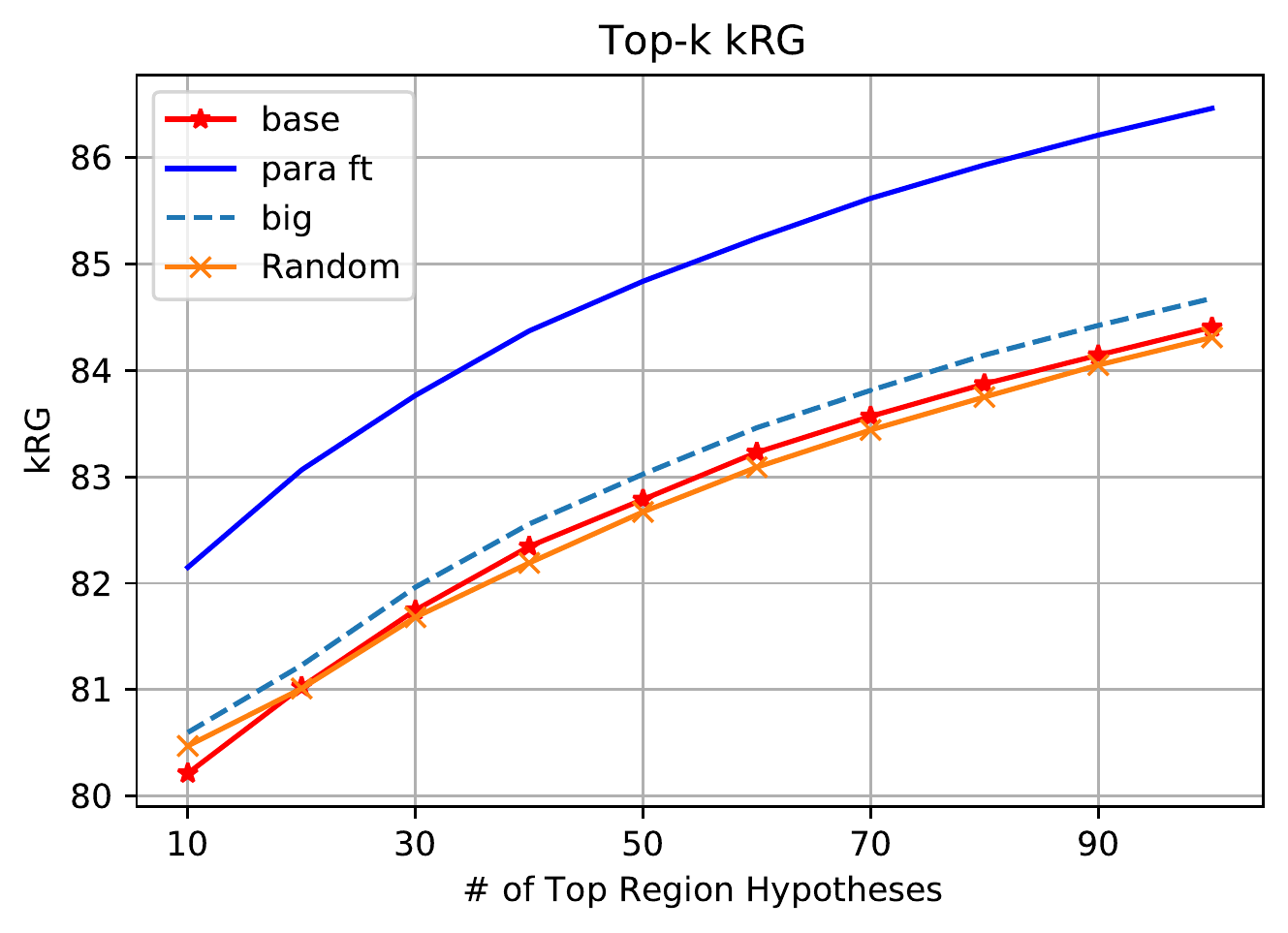}
}

\subfloat{%
    \includegraphics[width=0.9\linewidth]{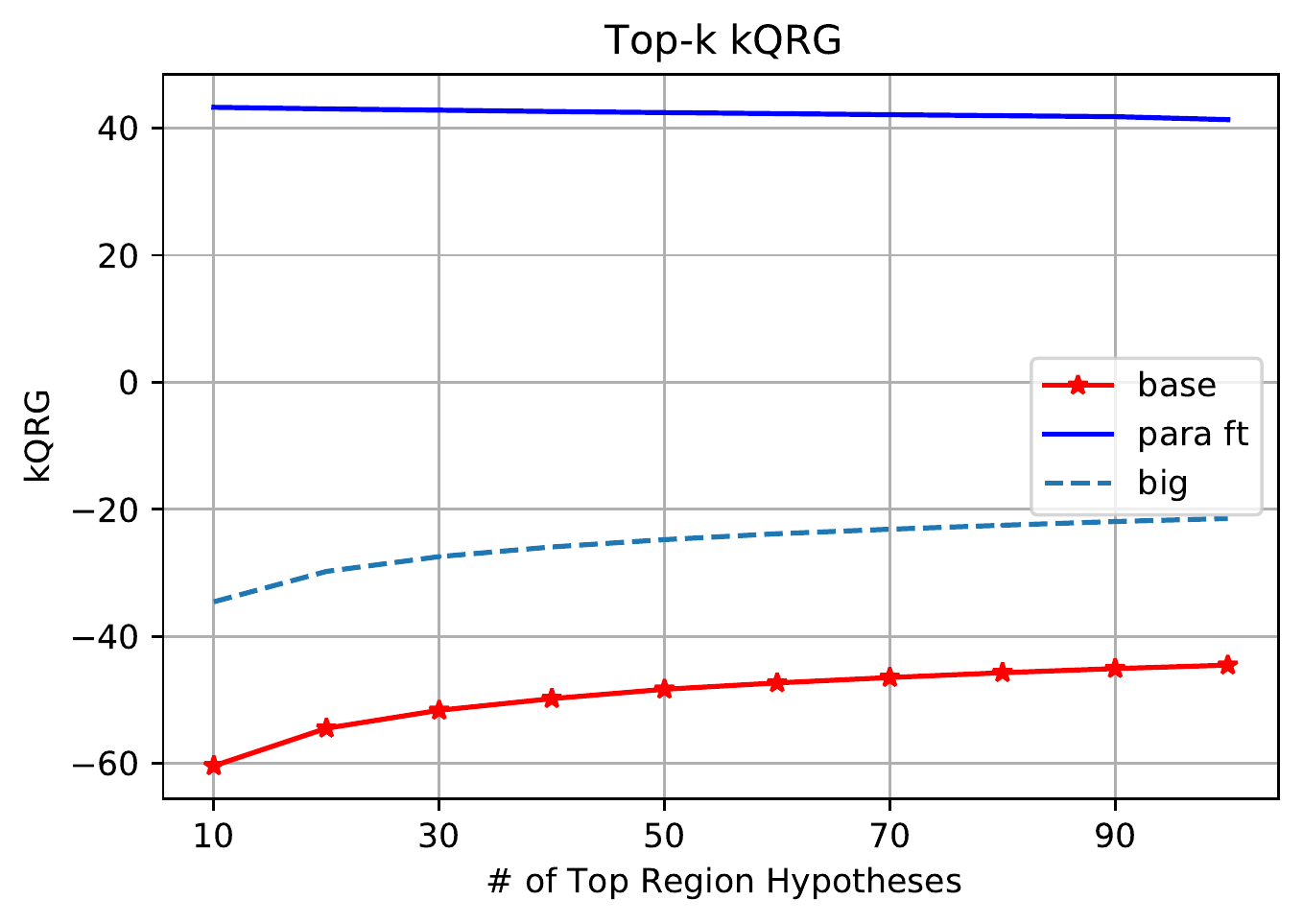}
}

\caption{kRG and kQRG for Transformer base, paraft and big models, with top-$k$ varies in \{10, 20, 30, 50, 75, 100\}}
\label{fig:different_k}
\end{figure}

We first report computational costs with different values of $k$, shown in Table \ref{table:increase_k}. 
The computational time and the number of expansions grow as $k$ increases.
When we enlarge the number of $k$ from 5 to 10, the time costs grow by about 1.9 times (\nicefrac{15,916.2 }{ 8,914.4}), which denotes an almost linear time cost with regard to $k$.  
Compared to \citep{stahlberg-byrne-2019-nmt}, our algorithms are more efficient -- Our top-$5$ algorithm operates two times of expansions and performs comparably with their algorithms in terms of computational time. 

{
Then, regarding the performance with different top-$k$, we plot models' $\text{kRG}$ and $\text{kQRG}$ with their top [10, 100] outputs. In Figure \ref{fig:different_k}, when we increase $k$, $\text{kRG}$ values of Transformer-Base (`base') and Transformer-Big (`big') stay close to the random permutation results, while the model trained with forward translation (`paraft') achieves a considerable gap over the random. 
The gap remains stable with larger values of $k$. 
The $\text{kQRG}$ values of all three models show good discrimination. We do not observe a trend of changing relative orders.
 
These results prove one important and favorable characteristic of our evaluation: \textbf{Both of our metrics are not sensitive to the choice of $k$}, which validates the usage with a lower value of $k$ to evaluate the model's distribution. }

In the main content of our paper, we mainly use top-$10$ results for our evaluation method for the trade-off between efficiency and effectiveness. 



\begin{table*}
\centering
\setlength\tabcolsep{2pt}
\begin{tabular}{l||c | c}
\toprule
\bf Method & \bf Time Cost (seconds) & \bf Num Expansions\\
\midrule
\citet{stahlberg-byrne-2019-nmt} & 8,064.0 & 2,769.6 \\
\midrule
Exact Top-$5$ w/ BS lower bounds & 8,914.4 & 6,029.4 \\
Exact Top-$10$ w/ BS lower bounds & 15,916.2 & 10,865.9 \\
Exact Top-$20$ w/ BS lower bounds & 28,313.9 & 19,155.8 \\
\bottomrule
\end{tabular}
\caption{Computational time and expansions for exact search algorithms when $k$ increases.}
\label{table:increase_k}
\end{table*}

\begin{table*}[t]
\centering
\setlength\tabcolsep{2pt}
\begin{tabular}{c||c |r| l}
\toprule
\bf Rank & \bf LogProb & \bf BLEU & \bf hypothesis \\
\midrule
Ref & - & 100.00 & \makecell[l]{\textbf{Zwei Anlagen so nah beieinander: Absicht oder} \\ \textbf{Schildbürgerstreich? <EOS>}} \\
\midrule
1 & -9.04 & 00.00 & \makecell[l]{<EOS> \\ ~~} \\
\hline
2 & -10.13 & 20.45 &\makecell[l]{Zwei Leuchten so nah beieinander: absichtlich oder einfach \\ nur ein dummer Fehler? <EOS>} \\
\hline
3 & -10.40 & 07.47 & \makecell[l]{Zwei Leuchten so nahe beieinander: absichtlich oder einfach \\ nur ein dummer Fehler? <EOS>} \\
\hline
4 & -10.56 & 22.24 & \makecell[l]{Zwei Leuchten so nah beieinander: absichtlich oder nur \\ein dummer Fehler? <EOS>} \\
\hline
5 & -10.92 & 08.13& \makecell[l]{Zwei Leuchten so nahe beieinander: absichtlich oder nur \\ein dummer Fehler? <EOS>} \\
\hline
6 & -10.94 & 05.89 & \makecell[l]{Zwei Leuchten so nahe beieinander? <EOS>\\ ~~ } \\
\hline
7 & -11.10 & 22.24 & \makecell[l]{Zwei Leuchten so nah beieinander: absichtlich oder einfach \\ ein dummer Fehler? <EOS>} \\
\hline
8 & -11.15 & 37.60 & \makecell[l]{Zwei Leuchten so nah beieinander: Absicht oder einfach nur \\ ein dummer Fehler? <EOS>} \\
\hline
9 &-11.21 & 17.63 & \makecell[l]{Zwei Leuchten so nah beieinander? <EOS>\\ ~~} \\
\hline
10 & -11.39 & 40.90 & \makecell[l]{Zwei Leuchten so nah beieinander: Absicht oder nur ein \\ dummer Fehler? <EOS>} \\
\bottomrule
\end{tabular}
\caption{The generated translations with top-$10$ decoding. The source sentence is \emph{"Two sets of lights so close to one another: intentional or just a silly error?"}}
\label{table:case_study}
\end{table*}

\section{Case Study}
This section provides a case study for English-German translation outputs for our Exact Top-$k$ decoding algorithm. Table \ref{table:case_study} shows the generated hypotheses, their corresponding log probabilities, and BLEU scores.

There are several problems of models' generated outputs based on the example: 
First, the ranking problem we argue in the main content apparently exists, which is demonstrated in our provided example. For instance, the model gives the highest score to an empty hypothesis (only <EOS>), which ranks the model's mode hypothesis the worst in the hypothesis space. 
Second, the model ranks some sub-optimal hypotheses in the top-$10$ rankings, like 2-nd, 4-th, 7-th, 10-th. 
However, the best hypothesis is ranked only at the 10-th position. 
It can also prove the existence of the ranking problem.
Third, the model favors shorter hypotheses. The hypotheses at rank positions 1-st, 6-th, and 9-th are much shorter than the others. The short hypotheses have roughly similar scores compared with the longer ones. Furthermore, most of the hypotheses share a similar prefix, which is similar to the reference, demonstrating that the model can find proper translations with incorrect log probabilities.
Those problems indicate the existence of an under-confidence problem, which is in line with our findings in Section \ref{sec:find_techinique}.

{
\section{Limitations}
We summarize our proposed method has two limitations. 
First, each of our approximations has its own limitations. Speaking of top region, the proposed exact search algorithm is computational extensive and local, meaning that it may be limited by its representativeness of the hypothesis space. As for Monte Carlo sampling, the evaluation is fast and more global but captures only coarse-grained model errors.
Even so, our two approximations can complement each other's limitations.
Second, our proposed metrics are dependent with the value of $k$ and choice of translation function. Specifically, for kRG, when we increase $k$ (Figure \ref{fig:krg_vs_k}), the random result also increases. For kQRG, we use COMET in our main content and report ChrF results in Appendix. These two results have very different scale and upper/lower bounds. This may lead to difficulty in interpretation.
}




\end{document}